\documentclass{article}
\usepackage[preprint]{neurips_2026}

\usepackage[utf8]{inputenc}
\usepackage[T1]{fontenc}
\usepackage{hyperref}
\usepackage{url}
\usepackage{booktabs}
\usepackage{amsmath}
\usepackage{amssymb}
\usepackage{microtype}
\usepackage{graphicx}
\usepackage{subcaption}
\usepackage{xcolor}
\usepackage{natbib}
\usepackage[most]{tcolorbox}
\usepackage{bm}

\definecolor{colGen0}{HTML}{555555}         
\definecolor{colUnfiltered}{HTML}{B37A00}   
\definecolor{colHK}{HTML}{009E73}           
\definecolor{bgGen0}{HTML}{F2F2F2}
\definecolor{bgUnfiltered}{HTML}{FEF7E0}
\definecolor{bgHK}{HTML}{E0F5EE}
\definecolor{frameUnfiltered}{HTML}{E69F00}
\definecolor{frameHK}{HTML}{009E73}

\tcbset{
  examplebox/.style n args={3}{
    enhanced, breakable,
    boxrule=0.6pt,
    colframe=#2,
    colback=#3,
    colbacktitle=#1,
    coltitle=white,
    fonttitle=\small\bfseries,
    toptitle=3pt, bottomtitle=3pt,
    left=6pt, right=6pt, top=4pt, bottom=4pt,
    fontupper=\small,
    sharp corners,
  }
}

\newcommand{\hk}{\hat{H}_{K}}
\newcommand{\hl}{H_{L}}
\newcommand{\hx}{\hat{h}_{\times}}

\title{No Model Required: Text Entropy Rate Filtering\\
  Mitigates Iterative Fine-Tuning Collapse}

\author{%
  Lewis Mitchell \\
  Adelaide Data Science Centre \\
  School of Mathematical Sciences \\
  Adelaide University \\
  Adelaide SA 5005, Australia \\
  \texttt{lewis.mitchell@adelaide.edu.au}
}

\begin{document}
\maketitle

\begin{abstract}
Iterative fine-tuning on synthetic data causes \emph{model collapse}: output
diversity narrows as rare patterns are progressively lost, a signature most
visible as phrase-level repetition. Existing
mitigations either require model log-probabilities, an external oracle, or continued
access to real human data. Here we develop a new approach grounded in mathematical information theory: the non-parametric Kontoyiannis entropy rate estimator
$\hk$, computed entirely from raw text via match-length statistics, with no
model of any kind. We show that this is in fact a \emph{superior} training-data filter on text-diversity
metrics in a fully-synthetic, single-lineage fine-tuning setting.
In a six-generation QLoRA collapse experiment on
Llama-3.1-8B, logprob-based filtering (the most established model-access-requiring baseline) provides
no significant text-diversity benefit on any metric ($p > 0.23$), whereas
$\hk$-filtering yields $+42\%$ unique trigrams, $+30\%$ vocabulary, and
$-19\%$ repetition (all $p < 0.001$). We validate $\hk$ as a cross-domain
entropy proxy ($\beta = 0.924$, $R^2 = 0.746$) and collapse detector
($\rho = +0.454$, $p < 0.0001$) across 4~domains, 2~temperatures,
2~generator--scorer model pairs, and 1{,}520 generated documents.
Our results demonstrate that information theoretic approaches to collapse mitigation are efficient, and suggest new approaches for maintaining multi-agent diversity.
\end{abstract}

\section{Introduction}
\label{sec:intro}

Large language models trained on their own outputs face a self-reinforcing
degradation known as \emph{model collapse}: when a model's outputs are
used to fine-tune successor models, the entropy of the output distribution
declines across generations, diversity narrows, and rare linguistic patterns
are progressively lost~\citep{shumailov2023,alemohammad2023}. As synthetic content becomes an increasing fraction of web-scraped training
corpora, collapse mitigation is shifting from a theoretical concern to an
engineering requirement for production training pipelines.

Existing mitigations for model collapse fall into one of three categories: 
(A)~model
logprob filters such as \emph{surplexity filtering}~\citep{gambetta2024},
which scores documents by the model's own per-token Shannon entropy and retains
those the model finds surprising; 
(B)~external oracle or classifier
approaches utilising an independent language model; and 
(C)~data accumulation strategies that re-inject real human data
at each generation~\citep{gerstgrasser2024,bertrand2024}. 
Among filtering approaches that do not require the injection of real, expensive human data, surplexity filtering is the most established model-access-requiring baseline.
\citet{gambetta2024} demonstrate that it delays collapse as measured
by the model's own entropy and perplexity scores, but its effect on
independent text-diversity measures has not been evaluated. Throughout this
paper we compare against mean per-token Shannon entropy $\hl$ specifically,
which is closely related to but not necessarily identical to surplexity:
both are monotonically increasing functions of the model's output entropy,
but surplexity uses the sample surprise of the generated tokens whereas
$\hl$ averages over the full output distribution, so the two induce highly
similar but not necessarily identical document rankings. Moreover,
\citet{guo2024decline} find that linguistic acceptability filtering can
actually \emph{worsen} collapse, underscoring that the choice of filtering
criterion is consequential.

Surplexity filtering requires a forward pass through the model being trained,
with access to the full next-token distribution at each position. Such access
is unavailable for closed-source APIs, computationally expensive at scale, and
version-dependent -- the same document receives different surplexity scores as
the model is updated. Data accumulation avoids model access but presupposes a
continuing supply of human-authored text, which is costly and slow to generate, and unavailable in a fully
synthetic closed loop. A text-only collapse filter that requires no model of
any kind would be strictly preferable, even if it were found to be less effective than surplexity filtering.

Motivated by results from mathematical information theory, we propose the \textbf{Kontoyiannis entropy rate estimator}
$\hk$~\citep{kontoyiannis1998} as a text-only collapse filter and show that it
outperforms logprob-based filtering. $\hk$ estimates the entropy rate of a
symbolic sequence by measuring match lengths: at each position, it records the
length of the longest prefix of the remaining text that has previously appeared
in the document. Long matches signal repetition in the text; short matches signal
novelty. The Kontoyiannis estimator has previously been used to quantify information flow in social
media text~\citep{bagrow2019information,bagrow2018quoter,pond2020complex} and requires no
model, no API, and no GPU. To our knowledge, no prior work has applied
$\hk$ or any compression-rate statistic as a collapse filter; it is the only
proposed filter of this type requiring no model of any kind -- not a forward
pass, not an oracle, not a reference classifier.

\begin{figure}[t]
  \centering
  \includegraphics[width=\textwidth]{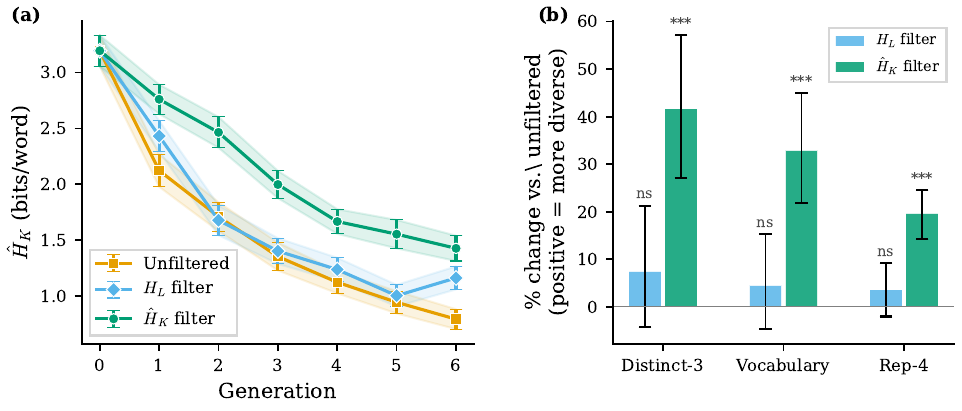}
  \caption{%
    Filtering on text-only entropy preserves diversity; logprob-based filtering
    does not. \textbf{(a)}~Kontoyiannis entropy rate $\hk$ across six
    fine-tuning generations of Llama-3.1-8B-Instruct under three training-data
    conditions: unfiltered (rapid collapse), $\hl$-filtered (intermediate), and
    $\hk$-filtered (slowest decline). Shaded bands: $\pm 1$~SEM ($n=80$
    documents per generation). \textbf{(b)}~Percentage change relative to the
    unfiltered baseline at generation~6 for three text-diversity metrics
    (Distinct-3, Vocabulary size, Rep-4). Error bars: bootstrap 95\% CIs.
    $\hk$-based filtering yields highly significant improvements on all three
    metrics (${}^{***}p<0.001$), whereas $\hl$-based filtering is
    non-significant on all three.
  }
  \label{fig:teaser}
\end{figure}

Figure~\ref{fig:teaser} previews our central result. In a six-generation
QLoRA fine-tuning experiment on Llama-3.1-8B-Instruct~\citep{dubey2024llama3},
each generation's model is fine-tuned from the previous one -- a continuous
model lineage reflecting the realistic deployed scenario -- and the filter
operates in a fully synthetic closed loop with no human data available for
re-injection. 
$\hl$-based training-data filtering -- the most established
model-access-requiring baseline -- provides no significant text-diversity benefit on any measured metric ($p>0.23$),
whereas \textbf{$\hk$-based filtering yields highly significant improvements} ($p<0.001$)
on Distinct-3, vocabulary size, and repetition rate -- despite $\hk$ requiring
no model access at all.
The two filters select substantially different training
documents (mean within-domain Jaccard similarity $\approx 0.35$), indicating
that $\hk$ is not merely approximating surplexity but operating on a complementary
axis: resistance to phrase-level repetition, the primary surface manifestation
of fine-tuning collapse.

We validate $\hk$ through three experiments: establishing it as a proxy for
logprob entropy across domains, temperatures, and model pairs
(Section~\ref{sec:proxy}); characterising when it does and does not track
collapse (Section~\ref{sec:collapse}); and demonstrating that $\hk$-based
filtering outperforms logprob-based filtering in a controlled iterative
fine-tuning experiment (Section~\ref{sec:phase5}).
We release our code and data for reproducibility\footnote{\url{https://github.com/lewismath/no-model-required}}.

\section{Background and Related Work}
\label{sec:background}

\paragraph{Model collapse.}
\citet{shumailov2023} demonstrate that iterative training on model-generated
data causes progressive loss of distributional tails: output variance shrinks
across generations as the model reinforces its existing biases. The phenomenon
is formalised by~\citet{alemohammad2023}, who show that without real-data
anchoring, iterative self-distillation inevitably converges to low-entropy
output. \citet{briesch2023} provide the first LLM-specific empirical study of
self-consuming training loops, finding that diversity declines in proportion to
the synthetic data fraction. Measuring collapse in terms of \emph{text-level}
diversity is a more recent development: \citet{guo2024decline} conduct
iterative fine-tuning on NLG tasks and track lexical, syntactic, and semantic
diversity across generations, observing that syntactic diversity deteriorates
fastest and creative tasks are most vulnerable. Both works characterise
collapse through output distribution statistics; we measure it directly via the
entropy rate of the generated text. Analogous diversity collapse has been
documented in multi-agent LLM systems, where structural coupling during
iterative interaction drives premature semantic convergence even without any
weight update~\citep{chen2026diversity}; the present work identifies a
complementary mechanism operating through the fine-tuning data loop.

\paragraph{Mitigating model collapse.}
Existing mitigations fall into three broad categories, each imposing a
distinct requirement on the practitioner.

\emph{(A)~Model-logprob methods.}
\citet{gambetta2024} propose \emph{surplexity} -- the product of token-level
surprise and perplexity -- as a data-quality signal for collapse mitigation.
In practice, retaining high-surplexity documents is closely related to
filtering by the mean Shannon entropy of the generator's next-token
distribution, which we denote $\hl$. This is the text-level analogue of
perplexity-based data filtering widely used in large-scale pretraining
pipelines~\citep{wenzek2020ccnet}. \citet{zhu2024synthesize} resample tokens
whose predicted probability exceeds a threshold from a prior language model,
requiring token-level logprobs during generation, while
\citet{alemohammad2023} show that biased or top-$K$ sampling -- retaining only
the highest-scoring synthetic samples each generation -- can slow collapse when
a quality score derived from model internals is available. All three methods
require access to the generating model's output distribution.

\emph{(B)~External-oracle methods.}
\citet{feng2025} argue that scaling up with synthesised data requires an
external verifier -- a stronger model or human annotator -- to score and filter
synthetic samples. \citet{yi2025} extend this line with convergence
guarantees for verifier-based filtering, but identify a critical caveat:
long-run convergence is toward the \emph{verifier's} biased knowledge centre,
not the true data distribution, unless the verifier is perfect. This finding
motivates the search for estimators that introduce no external model bias.

\emph{(C)~Real-data accumulation.}
\citet{gerstgrasser2024} prove that accumulating all prior real and synthetic
data yields a finite error bound and validate this empirically.
\citet{bertrand2024} show that retaining a fixed fraction of real data each
generation guarantees stability, and \citet{fu2025} establish that even a
constant proportion of real data ensures convergence in transformers. All
three approaches require continued access to human-authored data at each
training round.

Each type of mitigation thus requires either model internals~(A), a stronger external
model~(B), or curated human data~(C), imposing an extra burden to mitigate model collapse regardless of the approach taken.
In this work we introduce a novel filter in
category~(D): $\hk$ is computed from raw text alone -- no model, no API, no
GPU, no reference corpus -- making it, to our knowledge, the only completely
model-free and reference-free model collapse filter proposed to date.

\paragraph{Relationship to data deduplication.}
Min-hash near-deduplication~\citep{lee2022deduplicating,penedo2024fineweb}
is the standard model-free text-quality filter in large-scale pretraining
pipelines, and is the one existing method that, like $\hk$-filtering, requires
neither model access nor an external verifier. It removes near-duplicate
\emph{documents} from the training corpus (a between-document operation),
whereas $\hk$-filtering selects documents by within-document phrase variety
(a within-document operation). The two approaches are complementary:
deduplication prevents the corpus from containing many copies of the same
text, while $\hk$-filtering ensures each document has not collapsed into
internally repetitive structure. A combined pipeline (deduplicate first,
then $\hk$-filter) would address both failure modes.

\paragraph{The Kontoyiannis estimator.}
\citet{kontoyiannis1998} prove that the estimator
\begin{equation}
  \hk = \frac{n \log_2 n}{\sum_{i=1}^{n} \Lambda_i}
  \label{eq:hk}
\end{equation}
converges almost surely to the entropy rate $H$ of any stationary ergodic
process -- no parametric assumptions on the token distribution are
required -- where $\Lambda_i$ is the length of the shortest prefix of
$x_{i}, \ldots, x_n$ that does not appear as a contiguous substring in
$x_1, \ldots, x_{i-1}$; near the end of a document, where no such
non-appearing prefix may exist, $\Lambda_i$ is bounded by the remaining
sequence length $n-i$, so the sum in Equation~\eqref{eq:hk} remains
well-defined without special-casing. Reported $\hk$ values use the
\texttt{ProcessEntropy} package's tokenizer\footnote{%
  \url{https://github.com/tobinsouth/ProcessEntropy}}:
an NLTK TweetTokenizer pass, lower-cased, with non-alphanumeric characters
stripped from within each token~\citep{south2022information}. Real-time filtering decisions during
fine-tuning (Section~\ref{sec:phase5}) instead use a faster whitespace
tokenization (lower-case; non-alphanumeric characters replaced with
whitespace) for computational efficiency; the two agree closely in
practice (Spearman $\rho=0.997$ on a held-out sample), so the choice
does not affect which documents are selected.
The estimator's consistency is grounded in the Shannon--McMillan--Breiman
theorem~\citep{cover2006elements}: for a stationary ergodic source,
$-\frac{1}{n}\log_2 P(X_1,\ldots,X_n)\to H$ almost surely, and
$\Lambda_i$ serves as a nonparametric proxy for local self-information at
position $i$ -- long matches signal low surprise; short matches signal high
surprise.
The match lengths also connect $\hk$ to Lempel--Ziv (LZ) data
compression~\citep{ziv1977universal,ziv1978compression}: each $\Lambda_i$
identifies the shortest phrase at position $i$ not resolvable from prior
context -- the elementary step of LZ parsing -- so $\hk$ is interpretable
as a normalised LZ compression rate. While compression-based statistics have
been used for general data quality assessment, to our knowledge no prior work
has applied $\hk$ or any LZ-derived metric as a model collapse mitigation filter.
At finite $n$, $\hk$ is upward-biased: early positions have little past
context, so $\Lambda_i$ is artificially small and the denominator is
deflated.
We control for this by computing $\hk$ on the first 1{,}500 word tokens of
each document and discarding shorter texts (see Section~\ref{sec:proxy});
the residual finite-sample bias is reported as a limitation in
Section~\ref{sec:discussion}.
The Kontoyiannis estimator provides a model-free measure of sequential complexity that
depends only on the text
itself~\citep{bagrow2019information,smart2022entropy}.

\paragraph{Fine-tuning.}
We use QLoRA~\citep{dettmers2023qlora} -- 4-bit NF4 quantisation with
low-rank adaptation~\citep{hu2022lora} -- for all fine-tuning experiments.
QLoRA makes multi-generation fine-tuning tractable on a single A100 GPU,
enabling the iterative collapse and filtering experiments described in Sections~\ref{sec:collapse} and~\ref{sec:phase5}.

\section{$\hk$ as a Proxy for Logprob Entropy}
\label{sec:proxy}

We first ask whether $\hk$ is a suitable proxy for $\hl$, by comparing the two metrics on a set of synthetic documents.

\subsection{Cross-domain and cross-temperature validation}

We generated 200 documents using GPT-4o across 4~domains (encyclopedic,
creative, scientific, conversational), 2~temperatures ($T \in \{0.5,
1.0\}$; higher temperatures were excluded in preliminary testing because too
large a fraction of sampled tokens fell outside the top-20 returned
logprobs), and 25~topics per domain; 197 documents passed a reliability
filter requiring that fewer than 5\% of a document's tokens have
unrecoverable (invalid) logprob coverage.
For each document we computed $\hk$ via Equation~\eqref{eq:hk} applied to the
first 1{,}500 word tokens, and $\hl$, the mean per-token Shannon entropy of
the generator's next-token distribution, renormalised over the top-20 returned
logprobs.

An OLS regression of $\hk$ on $\hl$ with domain fixed effects yields
$\beta = 0.924$ ($R^2 = 0.746$, $p < 0.0001$; Figure~\ref{fig:proxy}a shows the $T=1.0$ subset).
A likelihood-ratio test for domain-specific slopes fails to reject equality
($p = 0.41$), confirming a \emph{domain-invariant slope}: the relationship
between $\hk$ and $\hl$ is consistent across domains up to a domain-level
intercept offset. Domains differ in baseline entropy -- creative and conversational text is
more varied; scientific text more constrained -- but $\hk$ scales with $\hl$ at
the same rate in each.

The domain-invariant slope has a practical consequence for data filtering.
Because filtering requires only the \emph{rank order} of documents by entropy
(selecting the top-$k$), not absolute calibration, a single $\hk$ computation
suffices for within-domain ranking. In a mixed-domain corpus, domain
stratification remains necessary to prevent over-selection from
naturally higher-entropy domains.

\subsection{Cross-model validation}

\begin{figure}[t]
  \centering
  \begin{subfigure}[b]{0.32\textwidth}
    \includegraphics[width=\textwidth]{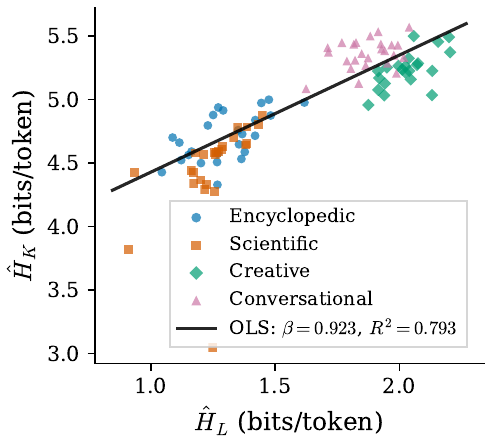}
    \caption{Cross-domain validation: GPT-4o corpus ($T=1.0$, $n=97$).}
  \end{subfigure}
  \hfill
  \begin{subfigure}[b]{0.65\textwidth}
    \includegraphics[width=\textwidth]{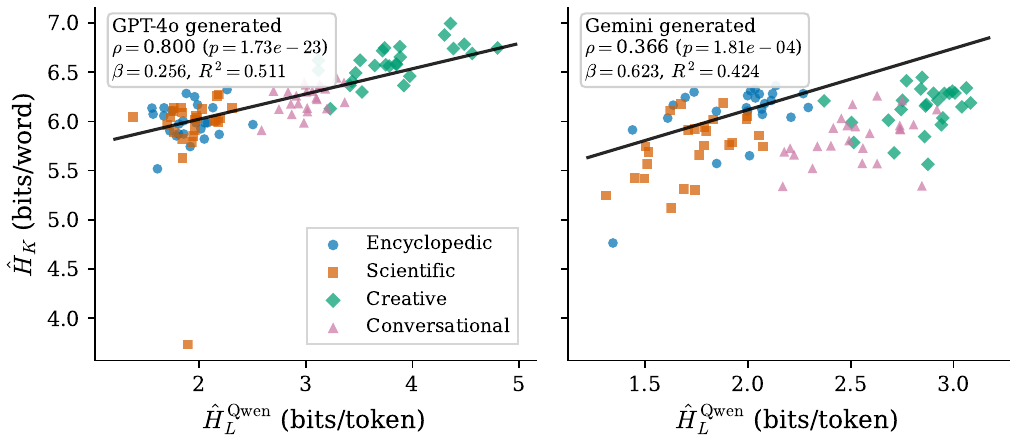}
    \caption{Cross-model validation: GPT-4o and Gemini corpora, both scored by
      Qwen2.5-7B-Instruct.}
  \end{subfigure}
  \caption{%
    $\hk$ versus $\hl$ across domains and model pairs. Points are coloured by
    domain. The slope is domain-invariant in both panels: $\hk$ scales with
    $\hl$ at a consistent rate regardless of domain, generating model, or
    scoring model.
  }
  \label{fig:proxy}
\end{figure}

The cross-domain experiments above use the same model (GPT-4o) for both
generation and $\hl$ scoring.
Here we test whether the $\hk$/$\hl$ relationship holds when $\hl$ is
computed by a different model family altogether.
We generated additional corpora using Gemini~1.5 Pro~\citep{team2024gemini} and
scored all documents -- both GPT-4o-generated and Gemini-generated -- using
Qwen2.5-7B-Instruct~\citep{yang2024qwen25} as an independent scorer.
Both the GPT-4o/Qwen and Gemini/Qwen pairings reproduce the positive
$\hk$/$\hl$ relationship (Figure~\ref{fig:proxy}b), confirming that $\hk$ measures a
property of the text itself, not of any particular model's distribution over
that text.

\subsection{Scope: proxy validity on collapse-experiment data}

The validations above use text generated under controlled conditions
(varied temperature, well-distributed topics). To assess whether the relationship
extends to the data used in the collapse experiment itself, we compute
$\hk$ and $\hl$ on the 80 generation-0 documents produced by
Llama-3.1-8B-Instruct (the base model, before any fine-tuning). The
correlation is near zero and non-significant ($r = 0.18$, $p = 0.11$,
$n = 80$). This is not a failure of the estimator but a known property of
the base model's output distribution: as documented in Figure~\ref{fig:supp_gen0},
the base Llama model produces a wide bimodal $\hl$ distribution (SD\,=\,2.17)
driven by topic-dependent familiarity rather than document-level diversity.
Within each topic, $\hl$ is nearly constant; between topics it varies by
up to 7 bits. $\hk$, which measures within-document phrase variety, is
orthogonal to this between-topic variation. The proxy relationship holds
when $\hl$ reflects genuine document-level entropy variation, but not when it
is dominated by topic-familiarity confounds -- and this is the root cause of
$\hl$-filtering's failure in Section~\ref{sec:phase5}.

\section{Detecting Model Collapse Without Logprobs}
\label{sec:collapse}

Having established $\hk$ as a cross-sectional proxy for $\hl$, we next ask
whether it tracks collapse \emph{over generations}. Under iterative rephrasing,
$\hk$ and $\hl$ co-decline at indistinguishable rates ($\delta = -0.004$,
$p = 0.59$), but within-generation rank concordance is near zero (Spearman
$\rho \approx 0$), as rephrasing homogenises surface statistics across
documents and eliminates the variance needed to discriminate between estimators. The fine-tuning regime is more informative, as we explore below.

\subsection{Iterative fine-tuning collapse}

Fine-tuning collapse operates through a different mechanism than rephrasing.
Rather than paraphrasing existing content, the model's \emph{generative
style} drifts: text exhibits increased phrase repetition, reduced vocabulary
diversity, and narrower syntactic range (see examples in Appendix \ref{app:examples}). 
We conducted a 6-generation QLoRA
fine-tuning experiment (Llama-3.1-8B-Instruct, unfiltered training data) and
computed within-topic $\hk$/$\hl$ concordance across generations for each of
39~topics (one topic was excluded due to insufficient within-generation
variance for reliable rank correlation).

\begin{figure}[t]
  \centering
  \begin{subfigure}[b]{0.62\textwidth}
    \includegraphics[width=\textwidth]{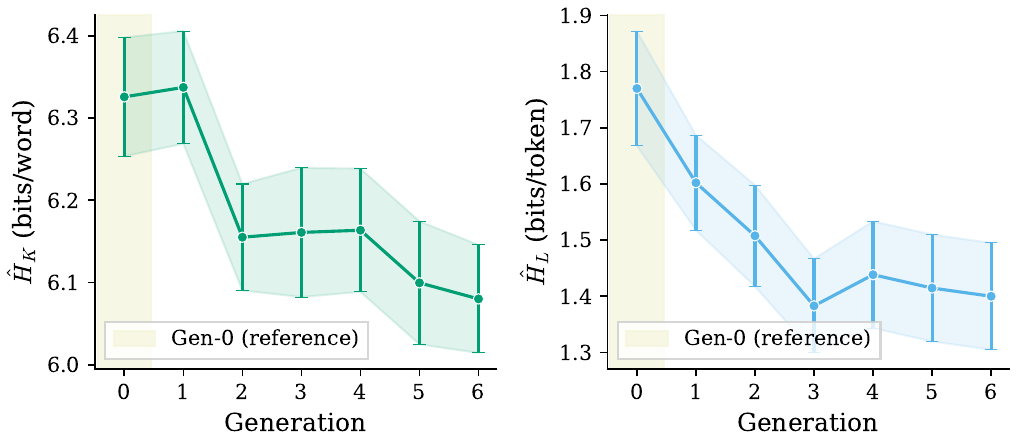}
    \caption{Iterative rephrasing collapse: mean $\pm$ SEM of $\hk$ and $\hl$ across rephrasing
      generations. Both estimators decline together ($\delta = -0.004$, $p = 0.59$).
      Shaded region: gen-0 reference (length-mismatched).}
  \end{subfigure}
  \hfill
  \begin{subfigure}[b]{0.35\textwidth}
    \includegraphics[width=\textwidth]{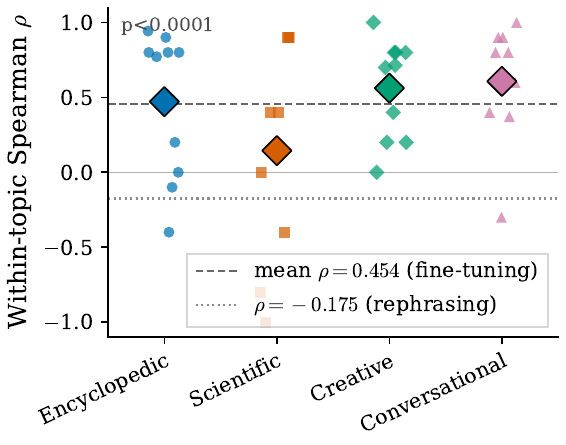}
    \caption{Iterative fine-tuning collapse: within-topic Spearman $\rho$ under fine-tuning collapse.
      Each point is one topic; diamonds show domain means.}
  \end{subfigure}
  \caption{%
    $\hk$ as a collapse detector under two regimes. Under rephrasing
    (a), $\hk$ and $\hl$ co-decline but within-generation concordance is $\approx 0$. Under fine-tuning (b), within-topic concordance is strongly positive
    ($\rho = +0.454$, $p < 0.0001$), driven by phrase-level repetition that
    $\hk$ directly measures.
  }
  \label{fig:collapse}
\end{figure}

Under fine-tuning collapse, within-topic Spearman concordance is
$\rho = +0.454$ ($p < 0.0001$), with 30 of 39~topics showing positive
concordance (Figure~\ref{fig:collapse}b). This contrasts sharply with the
rephrasing scenario ($\rho = -0.175$, non-significant). The explanation is
that fine-tuning collapse changes the surface statistics of generated text in
ways $\hk$ directly measures -- repetitive phrase structure, reduced $n$-gram
diversity -- whereas rephrasing preserves the within-generation homogeneity
that masks both estimators. This result establishes $\hk$ as a valid
\emph{detector} of fine-tuning collapse specifically, and motivates its use as
a filter in that regime.

\section{$\hk$ as a Training Data Filter}
\label{sec:phase5}

Having established that $\hk$ is a good collapse detector, we now consider our central question: can $\hk$ be used to filter training data and mitigate collapse; and, if so, does it outperform $\hl$, the most established model-access-requiring baseline?

\subsection{Experimental design}

We run three parallel 6-generation QLoRA fine-tuning chains on
Llama-3.1-8B-Instruct. At each generation, 80~documents are generated
across 4~domains (10~topics $\times$ 2~documents per topic per domain);
40~are selected for training according to the condition:
\begin{description}
  \item[\textnormal{Unfiltered}:] random selection (seed fixed per generation);
  \item[\textnormal{$\hl$-filter}:] top 40 by mean per-token Shannon entropy
    (logprob-based surplexity; see Section~\ref{sec:intro} for the
    distinction between $\hl$ and surplexity proper); and
  \item[\textnormal{$\hk$-filter}:] top 40 by Kontoyiannis entropy rate
    (text-only).
\end{description}
Selection is domain-stratified (10~documents per domain), so both filters
operate within domain. Generation-0 documents are drawn from the base model
and are shared across conditions. Training uses 4-bit NF4 QLoRA (rank~16,
$\alpha=16$, learning rate $2 \times 10^{-4}$, cosine schedule, 3~epochs per
generation, effective batch size~8). Documents selected for training are truncated at the onset of 4-gram
repetition; this preprocessing does not apply to generated outputs, which
are evaluated as-is (see Appendix~\ref{app:examples}). The experiment produces
1{,}520 documents in total: 80 shared generation-0 documents plus
3~conditions $\times$ 6~further generations $\times$ 80~documents.

\paragraph{Reproducibility.} Generation uses a fixed decoding temperature of
$1.0$ throughout. Prompts follow a fixed template with a domain-specific
system prompt (e.g., for encyclopedic: ``You are writing an encyclopaedia.
Write in a neutral, factual, Wikipedia-style tone.'') and a shared user
prompt, ``Write a detailed, long-form piece about the following topic. Write
at least 3{,}000 words. Do not stop early.\ Topic: \{topic\}.'' The topic set
(10 per domain $\times$ 4 domains) is identical across all six generations
and across all three conditions: the only thing that varies by condition is
which 40 of the 80 candidate documents are selected for training at each
generation and, from generation~1 onward, which fine-tuned checkpoint
generates the following round's candidate documents.

\subsection{Results}

Table~\ref{tab:phase5} summarises outcomes at generation~6 on four metrics.
$\hk$ (bits/word) and three text-diversity measures -- Distinct-3
\citep{li2016diversity}, vocabulary size, and intra-document 4-gram repetition
rate (Rep-4) -- are computed on the 80 held-out generation-6 documents
per condition. Diversity metrics are shown as percentage change relative to
the unfiltered condition.

\begin{table}[t]
  \caption{%
    Generation-6 outcomes under three training-data conditions. Diversity
    metrics show \% change relative to unfiltered. $p$-values from Welch
    $t$-tests. $n=80$ documents per condition.
  }
  \label{tab:phase5}
  \centering
  \small
  \begin{tabular}{lrrrccc}
    \toprule
    Metric & Unfilt. & $\hl$-filt. & $\hk$-filt.
           & \multicolumn{1}{c}{$p$\,($\hk$\,v.\,U)}
           & \multicolumn{1}{c}{$p$\,($\hl$\,v.\,U)}
           & \multicolumn{1}{c}{$p$\,($\hk$\,v.\,$\hl$)} \\
    \midrule
    $\hk$ (bits/word)
           & 0.79 & 1.16          & \textbf{1.43}
           & ${<}0.001$ & $0.008$ & ${<}0.001$ \\
    Distinct-3 (\%)
           &  --   & $+7$\,\% & \textbf{+42\,\%}
           & ${<}0.001$ & $0.23$ & ${<}0.001$ \\
    Vocabulary (\%)
           &  --   & $+4$\,\%  & \textbf{+30\,\%}
           & ${<}0.001$ & $0.31$ & ${<}0.001$ \\
    Rep-4 (\%)
           &  --   & $-3$\,\%  & \textbf{-19\,\%}
           & ${<}0.001$ & $0.27$ & ${<}0.001$ \\
    \bottomrule
  \end{tabular}
\end{table}

\paragraph{$\hk$-filtering slows collapse.}
At generation~6, $\hk$-filtering retains $1.43$~bits/word versus $0.79$ for
unfiltered ($p < 0.001$) and $1.16$ for $\hl$-filtering ($p < 0.001$). On all
three text-diversity metrics, $\hk$-filtering produces large, significant
improvements over unfiltered training: $+42\%$ more unique trigrams, $+30\%$
larger vocabulary, and $-19\%$ less intra-document repetition (all
$p < 0.001$). These gains are consistent across all four domains (domain-level trajectories
and full metric suites across generations are shown in Appendix~\ref{app:supp}).
Illustrative generation-6 documents from the unfiltered and $\hk$-filter conditions
are shown in Appendix~\ref{app:examples}.

\paragraph{$\hl$-filtering provides no text-diversity benefit.}
Despite having access to the model's log-probabilities, $\hl$-filtering
produces no significant improvement on any of the three text-diversity metrics
($p > 0.23$ on all). It does outperform unfiltered training on the $\hk$
metric itself ($p = 0.008$), but the effect is less than half that of the
$\hk$-filter. The failure has a clear mechanistic explanation: after even one fine-tuning
step, the $\hl$ distribution compresses dramatically (gen-0 SD~$= 2.17$;
gen-1 SD~$= 0.13$), collapsing the ranking signal the filter depends on.
Within a generation whose outputs are already near-uniform in $\hl$, the
filter degenerates to near-random selection.

\paragraph{Mixed-effects analysis.}
A mixed-effects regression ($\hk \sim \text{generation} \times
\text{condition} + (1|\text{topic})$, baseline: unfiltered) decomposes the
trajectory into a starting-quality component and a rate-of-collapse component.
$\hk$-filtering produces a significant intercept shift ($\beta = +0.37$~bits,
$p = 0.001$); because all conditions share the same generation-0 documents, this
reflects the shape of the fitted trajectories rather than a pre-existing quality
difference. The slope improvement is positive but does not reach significance
($\Delta\text{slope} = +0.058$~bits/gen, $p = 0.067$).
$\hl$-filtering produces neither a significant intercept shift nor a slope
improvement ($p > 0.1$ on both).

\paragraph{The filters select different documents.}
Within-domain Jaccard similarity between the $\hl$-filter and $\hk$-filter
selection sets has a mean of 0.35 (range 0.18--0.54 across generations and
domains). The two criteria do not approximate each other: $\hk$ selects on
phrase-level novelty, while $\hl$ selects on token-level model uncertainty.
The success of $\hk$-filtering despite this low overlap suggests that
phrase-level repetition, not token-level surprise, is the more informative
axis for collapse mitigation -- consistent with the observation that
fine-tuning collapse manifests primarily in surface repetition.

\paragraph{$\hk$ outperforms simpler text-only baselines.}
A natural objection is that any per-document diversity statistic might
replicate $\hk$'s advantage -- for instance, type--token ratio (TTR) or
bigram distinctiveness (Distinct-2). We assess this by simulating both
filters on the generation-0 pool, where all conditions share the same 80
documents and the comparison is exact. TTR@500-filtering selects training
documents with mean $\hk = 3.37$; Distinct-2-filtering selects mean
$\hk = 3.32$; $\hk$-filtering selects mean $\hk = 3.98$ -- a 18--20\%
advantage. Despite moderate positive correlations between TTR and $\hk$ in
the pool (Pearson $r = 0.54$), the selection sets overlap only 43\% by
Jaccard, indicating that $\hk$ recovers phrase-level structure that
per-document surface statistics do not capture. The difference is structural:
TTR and Distinct-2 aggregate over short local windows, whereas $\hk$ is
sensitive to long-range match structure -- precisely the property that
fine-tuning collapse erodes.
We repeated this static-pool comparison against two further model-free
criteria: an off-the-shelf gzip-compression ratio (least-compressible 40
documents selected per domain) and an explicit 4-gram repetition-rate filter
(lowest repeated-4-gram rate). Both correlate with $\hk$ only moderately
(gzip $r = +0.44$, $p < 0.001$; repetition rate $r = -0.36$, $p = 0.001$) --
comparable to the TTR correlation above -- and select substantially
different document sets (Jaccard $0.36$--$0.38$ against the $\hk$-filter
selection). As with the TTR/Distinct-2 comparison, this establishes that
$\hk$ is related to but not redundant with either criterion on the static
pool, but not downstream collapse-resistance, which we test directly by
running both as genuine six-generation training filters, replacing $\hk$ in
the filtering criterion but otherwise identical to Section~\ref{sec:phase5}.
At generation~6, $\hk$-filtering significantly outperforms both on every
metric ($\hk$, Distinct-3, Rep-4, vocabulary; all $p<0.005$, most
$p<10^{-4}$). The two static-pool-comparable criteria do not perform alike
as real filters, however: gzip-filtering gives a modest but genuine
diversity benefit over unfiltered training (Distinct-3 $+17\%$, vocabulary
$+17\%$, Rep-4 $-7\%$; all $p<0.03$), while repetition-rate filtering gives
none (all four metrics indistinguishable from unfiltered, $p>0.24$) despite
directly targeting repetition. $\hk$'s advantage over simpler model-free
heuristics is therefore not just a static-pool artefact: it persists, and
in repetition-rate filtering's case is entirely absent for the alternative,
under the same six-generation training loop that constitutes this paper's
central result.

\begin{figure}[t]
  \centering
  \includegraphics[width=\textwidth]{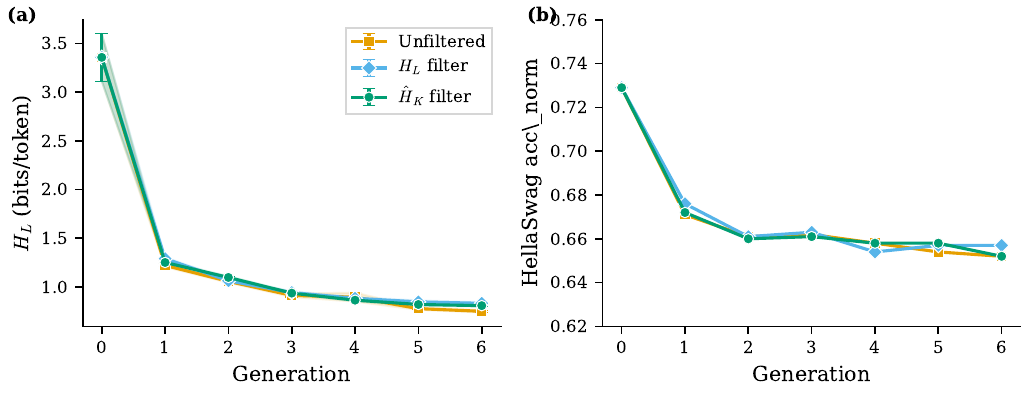}
  \caption{%
    \textbf{(a)}~$\hl$ (logprob entropy) trajectories across six fine-tuning
    generations for the three filtering conditions ($\pm 1$ SEM bands).
    All collapse to low $\hl$ by gen-1 and 
    track nearly identically, illustrating why the $\hl$~filter fails:
    post-collapse variance is too small to discriminate.
    \textbf{(b)}~HellaSwag normalised accuracy (Llama-3.1-8B-Instruct) across
    six generations. All three conditions degrade from $0.729$ (gen-0)
    to $\approx 0.652$--$0.657$ (generation~6) with no significant difference
    between conditions, indicating that text-diversity preservation does not
    translate to preserved general reasoning ability.
  }
  \label{fig:phase5_hl_hellaswag}
\end{figure}

\paragraph{External validation.}
We scored gen-6 documents from all conditions using
GPT-2~\citep{radford2019gpt2}, a 124M-parameter model with no relation to
training or generation pipelines. GPT-2 entropy shows the same directional ordering ($\hk$-filtered $>$
$\hl$-filtered $>$ unfiltered) but does not reach significance ($p = 0.11$),
limiting its value as independent validation; GPT-2's discriminative power on
long-form text is insufficient to detect the effect at the individual-document
level. We also evaluated each generation's fine-tuned checkpoint on
HellaSwag~\citep{zellers2019hellaswag} using Llama-3.1-8B-Instruct
(Figure~\ref{fig:phase5_hl_hellaswag}b). Accuracy drops sharply from
generation~0 to generation~1 ($0.729 \to \approx 0.672$) and then plateaus
near $0.652$--$0.657$ with no significant difference between conditions at any
generation. Thus $\hk$-filtering preserves text diversity but does not protect
general reasoning ability: the capability degradation introduced by iterative
fine-tuning is uniform across all three conditions. Two further downstream
benchmarks corroborate this: 5-shot MMLU and 8-shot GSM8K (1{,}000-example
subsamples each) on the same checkpoints show no significant difference
between conditions at generation~6 (MMLU: $67.6$--$67.6\%$; GSM8K:
$70.7$--$71.2\%$; all pairwise $\chi^2$ tests $p>0.8$), nor a significant
decline from generation~0 to generation~6 within any single condition
($p>0.15$) -- consistent with HellaSwag, general capability retention is
unaffected by which text-diversity filter is used, at this scale and over
this many generations.

\section{Discussion}
\label{sec:discussion}

\paragraph{Why $\hk$-filtering outperforms $\hl$-filtering.}
The Kontoyiannis estimator directly penalises repeated phrase structure: a
document with many recurring $n$-grams exhibits long match lengths and hence a
low $\hk$. Fine-tuning collapse manifests primarily as increased phrase
repetition, so $\hk$-filtering targets collapse's dominant surface signature
directly. $\hl$, by contrast, measures token-level uncertainty in the scorer
model -- a signal that is destroyed after a single fine-tuning step, as shown
in Section~\ref{sec:phase5}. At generation-0, $\hl$ primarily
reflects topic familiarity -- the wide bimodal distribution
documented in Figure~\ref{fig:supp_gen0} -- rather than document diversity, so
a criterion uncorrelated with document quality at generation-0 cannot select
high-quality training data regardless of downstream variance. The result
suggests that collapse is fundamentally a phrase-repetition phenomenon, and
that phrase-level statistics are better positioned to detect it than
token-level uncertainty. A literal reimplementation of surplexity confirms
this mechanistic prediction directly: its median absolute deviation across
the shared generation-0 pool is $4.40$, falling to $0.40$--$0.49$ (all three
conditions) after one fine-tuning step -- a $9$--$11$-fold narrowing,
comparable to $\hl$'s own collapse. (Generation-0 surplexity is heavy-tailed
-- a handful of unfamiliar-topic documents receive extreme values once
surprise is exponentiated into perplexity -- so median and MAD, not mean and
SD, are the meaningful summary here.)

\paragraph{Filter design determines direction of effect.}
Not all filters slow collapse: \citet{guo2024decline} find that linguistic
acceptability filtering (a RoBERTa classifier trained on CoLA)
\emph{accelerates} diversity decline, since such filters retain
conventional, high-frequency text and discard the inventive samples most
worth keeping. $\hk$-filtering operates on the opposite principle, selecting
for the phrase-level novelty that collapse erodes.

\paragraph{The Jaccard gap and Simpson's paradox.}
The reason why the two filters select such different documents
(Jaccard $\approx 0.35$) when $\hk$ and $\hl$ are well-correlated at the
corpus level ($R^2 = 0.746$) is a form of Simpson's paradox.
The \emph{between}-domain correlation is strong: creative and conversational
text have both high $\hl$ and high $\hk$; scientific text has both low. But
\emph{within} a domain and generation -- which is where filtering
operates, due to domain stratification -- the residual correlation is near
zero. The estimators measure genuinely different properties of
intra-domain text variation; $\hk$'s phrase-level
novelty  proves more informative for collapse resistance.

\paragraph{Practical implications.}
$\hk$ filtering requires no model access, no API calls, and no GPU: scoring a
1{,}500-word document takes under 10\,ms, and a pipeline producing $10^5$
candidate documents per round incurs negligible, trivially parallelisable
overhead. This makes collapse-resistant data curation practical where
logprob access is unavailable: closed-source API distillation,
cross-organisation data sharing, and federated training.

\paragraph{Limitations.}
Most importantly, $\hk$-filtering preserves text diversity but does not protect
general reasoning ability: HellaSwag accuracy degrades equally across all three
filtering conditions (Figure~\ref{fig:phase5_hl_hellaswag}b), and MMLU and
GSM8K show the same pattern, so our claim is scoped to distributional
fidelity of generated text, not downstream task performance. The filtering
advantage is also specific to fine-tuning collapse:
under iterative rephrasing, $\hk$ tracks the aggregate trajectory but cannot
distinguish among documents within a generation, and is therefore unsuitable as
a within-generation filter in that setting. The diversity gains measured here are lexical: MAUVE scores~\citep{pillutla2021mauve}
(GPT-2-Large features, first 512 tokens) are near-ceiling and uniform across
conditions at generation~6 ($0.987$--$0.993$), indicating embedding-space
similarity is preserved and $\hk$'s advantage is specific to within-document
match-length structure.
A separate concern is quality: selecting for match-length novelty could
favour incoherent or off-topic text over genuinely diverse text. We address
this directly with an LLM-judge pass (GPT-4o-mini, blind to condition)
scoring all 240 generation-6 documents (80 per condition) on two 1--5
scales, coherence and instruction-following, reading the full document so
that late-document degeneration is not missed. $\hk$-filtering scores
highest of the three conditions on both:
coherence $2.35$ versus $2.20$ for unfiltered ($p = 0.058$) and $2.29$ for
$\hl$-filtering; instruction-following $3.23$ versus $3.00$ for unfiltered
($p = 0.020$) and $3.15$ for $\hl$-filtering. If anything, then,
$\hk$-filtering's diversity advantage comes with a small quality benefit
rather than a coherence cost. Finally, our collapse experiment uses a single model family (Llama-3.1, 8B);
replicating with Llama-3.2-3B-Instruct (same corpus and protocol), $\hk$-filtering
again retains more diversity at generation~6 ($2.91$ bits/word versus $2.12$
unfiltered, $p<0.001$, and $2.41$ for $\hl$-filtering, $p=0.004$; $\hl$-filtering
again non-significant on all text-diversity metrics), with the $\hk$-vs-$\hl$
comparison itself reaching significance on the $\hk$ metric ($p=0.004$ vs
$p=0.088$ at 8B), providing stronger evidence the two filters are genuinely
different.
Beyond replication, the Kontoyiannis cross-entropy rate $\hx(T \mid S)$ is a
natural complement to $\hk$, selecting for documents that have diverged from
prior-generation content; the same signal could operationalise -- as an
automatic, text-only proxy -- the blind-writing independence phase
\citet{chen2026diversity} identify as the most effective structural
intervention against diversity collapse.

\paragraph{Broader impacts.}
The primary benefit is making collapse-resistant data curation accessible
without model access, reducing a practical barrier for organisations running
iterative fine-tuning under closed-source or resource-constrained conditions,
and counteracting homogenisation of AI-generated web content. The main negative
consequence is indirect: making fully-synthetic training pipelines more viable
may reduce the incentive to collect or compensate human-authored training data;
practitioners deploying such pipelines at scale should weigh this trade-off.

\section{Conclusion}
\label{sec:conclusion}

We have shown that the Kontoyiannis entropy rate estimator $\hk$ -- a
model-free, match-length measure of sequential complexity -- is an effective
training-data filter against fine-tuning-driven model collapse: at
generation~6 it produces $42\%$ more unique trigrams, $30\%$ larger
vocabulary, and $19\%$ less intra-document repetition than unfiltered
training, while logprob-based filtering shows no significant diversity
benefit. For iterative pipelines where model internals are inaccessible,
$\hk$-filtering offers a strong, zero-cost baseline.

\begin{ack}
This research was supported by the Australian Government through the Australian Research Council's Discovery Projects funding scheme (project DP210103700). Computational resources provided by the Phoenix HPC cluster at Adelaide University.
\end{ack}

\bibliographystyle{plainnat}
\bibliography{refs}

\newpage

\appendix

\section*{Appendices}

\section{Qualitative Examples of Fine-Tuning Collapse}
\label{app:examples}

The following excerpts illustrate the surface-level deterioration that $\hk$ measures.
Each example shows the same document topic at generation~0 (shared across all conditions)
and generation~6 under the unfiltered and $\hk$-filter conditions.
$\hk$ values are computed on the first 1{,}500 word tokens per document.

\subsection*{Example 1 (Conversational): \textit{Experiences with unexpected kindness from strangers}}

\begin{tcolorbox}[examplebox={colGen0}{colGen0}{bgGen0},
  title={Generation~0 \normalfont($\hk = 4.72$~bits/word)}]
\textbf{A Surprising Revelation: Unexpected Kindness from Strangers}

As I sit here, reflecting on the countless moments that have shaped my life, I find myself
pondering a peculiar yet profound truth. It's a truth that often goes unspoken, yet resounds
loudly in my mind: unexpected kindness from strangers can be the spark that ignites a chain
reaction of compassion, empathy, and connection. In this essay, I'll explore the unexpected
kindness that I've experienced firsthand, and I'll invite you to join me on a journey of
self-discovery, as we explore the transformative power of kindness from strangers.

\textbf{The Power of Random Acts of Kindness}

As I was walking home from work one day, lost in thought, I stumbled upon a small caf\'{e}.
I had been there before, but today, I spotted a poster advertising a free lunch for
low-income families. My heart ached as I watched a group of homeless individuals, struggling
to find any form of sustenance, pushing carts filled with fresh produce and leftovers.
[\ldots] As I took a bite, I watched a young boy, no more than 10 years old, sitting across
from me, staring intently at a drawing on the table. He seemed to be struggling to put the
pencil to paper, his eyes welling up with tears.
\end{tcolorbox}

\begin{tcolorbox}[examplebox={colUnfiltered}{frameUnfiltered}{bgUnfiltered},
  title={Generation~6, unfiltered \normalfont($\hk = 0.08$~bits/word)}]
\textbf{The Power of Unexpected Kindness: A Reflection on the Beauty of Human Connection}

As I sit here, reflecting on the past year, I find myself pondering the many unexpected
moments of kindness that have touched my life. It's a theme that has become increasingly
important to me, as I've come to realize the significance of the small acts of kindness that
can have a profound impact on our lives. In this essay, I'll share with you some of the most
memorable and heartwarming experiences I've had with unexpected kindness from strangers.
[\ldots]

\textit{[\ldots after eight early sections, the document collapses into verbatim-repeated paragraphs:]}

\textbf{The Unexpected Support System}

One of the most surprising experiences I've had with unexpected kindness was when I received
a message from a stranger, who asked me to join them for a support group. They invited me
to join a group of people who were going through similar struggles, and they promised that
they would be there to support me. I was hesitant at first, but I decided to take a chance.
I attended the support group, and it was a turning point for me. I felt a sense of connection
with the people around me, and I felt like I was not alone. [\ldots] I realized that
sometimes, the most unexpected gifts can come from the most unexpected places.

\textbf{The Unexpected Gift of Empathy}

One of the most surprising experiences I've had with unexpected kindness was when I received
a message from a stranger, who asked me to join them for a self-discovery retreat. They
invited me to join a group of people who were going through similar struggles, and they
promised that they would be there to support me. I was hesitant at first, but I decided to
take a chance. I attended the retreat, and it was a turning point for me. [\ldots]

\textit{[The above two paragraphs alternate and repeat verbatim approximately eight further times until the document length limit is reached.]}
\end{tcolorbox}

\begin{tcolorbox}[examplebox={colHK}{frameHK}{bgHK},
  title={Generation~6, $\hk$-filter \normalfont($\hk = 5.07$~bits/word)}]
\textbf{The Power of Unexpected Kindness: Lessons from Strangers}

I remember the day I stumbled upon a stranger who changed my life forever. It was a chilly
winter morning, and I was walking home from work, lost in thought. The cold air hit me like
a slap in the face, and I felt a surge of frustration and anger. [\ldots]

As I turned a corner, I noticed a woman standing on the sidewalk, her eyes downcast and her
hair disheveled. She was wearing a faded jacket and a pair of worn-out shoes, and she looked
like she had been walking for hours. [\ldots] She offered me the food, and I gratefully
accepted, feeling a sense of gratitude towards this stranger.

The woman, who I later learned was a young mother on her way to a job interview, smiled at
me as she handed me the food. ``Thank you,'' she said, her voice barely above a whisper.
``I'm so tired. I haven't had any breakfast in hours.'' I felt a surge of empathy for her,
and I wanted to help her in any way I could.
\end{tcolorbox}

Note that the $\hk$-filtered example ($\hk = 5.07$ bits/word) is a high-entropy
outlier relative to the condition's generation-6 mean ($\hk = 1.43$ bits/word);
it is selected here to illustrate the qualitative contrast, not to represent the
average filtered output.

The collapse in the unfiltered generation-6 document is qualitatively unmistakable: the
second half consists of near-verbatim repeated paragraphs that contribute no new content.
The $\hk$ metric reflects this directly -- long match lengths throughout the document's
latter portion drive the estimate to near zero. The $\hk$-filtered document opens with a
specific, situated narrative, consistent with its $\hk$ value exceeding generation~0 (the
filter selected an unusually high-entropy document at generation~0 to begin the lineage,
and the fine-tuned model preserved that stylistic register longer).

\subsection*{Example 2 (Scientific): \textit{Graphene oxide membranes for water desalination}}

\begin{tcolorbox}[examplebox={colGen0}{colGen0}{bgGen0},
  title={Generation~0 \normalfont($\hk = 1.65$~bits/word)}]
\textbf{Graphene Oxide Membranes for Water Desalination: A Review of their Mechanisms, Advantages, and Limitations}

Water desalination remains a pressing issue worldwide, particularly in regions with limited
access to freshwater resources. Traditional methods, such as reverse osmosis (RO), are
energy-intensive and often costly. [\ldots] Graphene oxide (GO) is a derivative of graphene,
where oxygen atoms are replaced by hydroxide groups on the graphene surface, enhancing
electrical conductivity, hydrolysis resistance, and water adsorption capacity.

\textbf{Mechanisms of GO Membranes}

The GO membrane is fabricated by depositing GO on a substrate, such as silicon dioxide
(SiO\textsubscript{2}) or carbon nitride, followed by deposition of a sacrificial layer such
as indium tin oxide (ITO). The GO layer is then cured and thermally treated to form a
composite membrane. The GO membrane exhibits unique properties: (1)~\textit{enhanced
hydrophilicity} -- surface energy significantly higher than pure graphene; (2)~selective ion
exclusion controlled by interlayer spacing [\ldots]
\end{tcolorbox}

\begin{tcolorbox}[examplebox={colUnfiltered}{frameUnfiltered}{bgUnfiltered},
  title={Generation~6, unfiltered \normalfont($\hk = 0.12$~bits/word)}]
\textbf{Graphene Oxide Membranes for Water Desalination: A Review of the Current State of the Art and Future Directions}

Graphene oxide membranes have emerged as a promising technology for desalination,
particularly for removing salt and other minerals from seawater and brackish water. Graphene,
a highly conductive and versatile two-dimensional material, has been extensively researched
and developed for various applications, including energy, electronics, and water treatment.
[\ldots] Desalination is the process of removing salt and other minerals from seawater and
brackish water to produce fresh water for human consumption, industry, and agriculture.
[\ldots] Graphene oxide membranes have been shown to be highly efficient in removing salt and
other minerals from water, making them an attractive alternative to traditional desalination
methods. [\ldots]
\end{tcolorbox}

\begin{tcolorbox}[examplebox={colHK}{frameHK}{bgHK},
  title={Generation~6, $\hk$-filter \normalfont($\hk = 1.19$~bits/word)}]
\textbf{Graphene Oxide Membranes for Water Desalination: A Review of the Current State of the Art and Future Directions}

Graphene oxide membranes have emerged as a promising technology for water desalination,
offering several advantages over traditional reverse osmosis (RO) and electrodialysis (ED)
techniques. [\ldots] The graphene oxide layer is typically 10--20~nanometres thick and
consists of a mixture of graphene oxide, hydroxyl groups, and other functional groups. [\ldots]

\textbf{Permeation Properties of Graphene Oxide Membranes}

Graphene oxide membranes have been shown to have exceptional permeation properties,
allowing for the separation of ions and molecules with high selectivity. The permeation
properties are influenced by the type of functional groups present, the thickness of the
membrane, and the temperature and pressure conditions. [\ldots]
\end{tcolorbox}

This example illustrates a different collapse mode from Example~1. Rather than verbatim
repetition, the unfiltered generation-6 document exhibits \emph{semantic dilution}: the
phrase ``removing salt and other minerals from seawater and brackish water'' appears three
times in the opening 200~words, while specific mechanistic details present in
generation~0 -- fabrication steps, substrate materials, numbered membrane properties -- are
absent. The $\hk$-filtered document partially recovers this specificity, retaining membrane
thickness values (10--20~nm), functional group details, and permeation mechanism structure.
Scientific-domain documents start with lower absolute $\hk$ values than conversational ones
(domain-level baseline effect, Section~\ref{sec:proxy}), so the absolute values are not
directly comparable across examples; the relative contrast within each example is the
relevant quantity.


\section{Supplementary Figures}
\label{app:supp}

\begin{figure}[ht]
  \centering
  \includegraphics[width=0.7\textwidth]{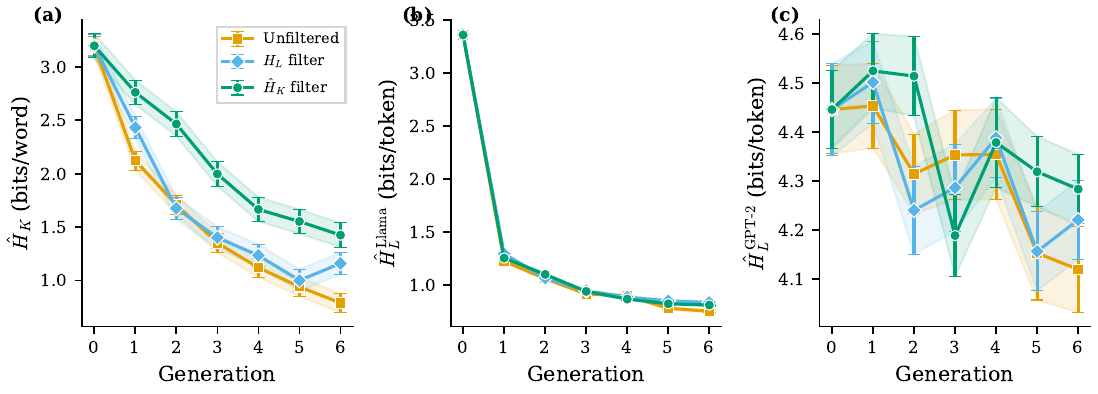}
  \caption{%
    Three entropy metrics across six fine-tuning generations:
    $\hk$ (text-only), $\hl^{\text{Llama}}$ (Llama-3.1-8B scorer),
    and $\hl^{\text{GPT-2}}$ (independent GPT-2 scorer). All three metrics
    show the same directional ordering at generation~6. The $\hl^{\text{Llama}}$
    panel illustrates why logprob-based filtering fails: after generation~0, the
    three conditions are nearly indistinguishable, leaving the $\hl$-filter
    little variance to exploit.
  }
  \label{fig:supp_three_metrics}
\end{figure}

\begin{figure}[ht]
  \centering
  \includegraphics[width=0.8\textwidth]{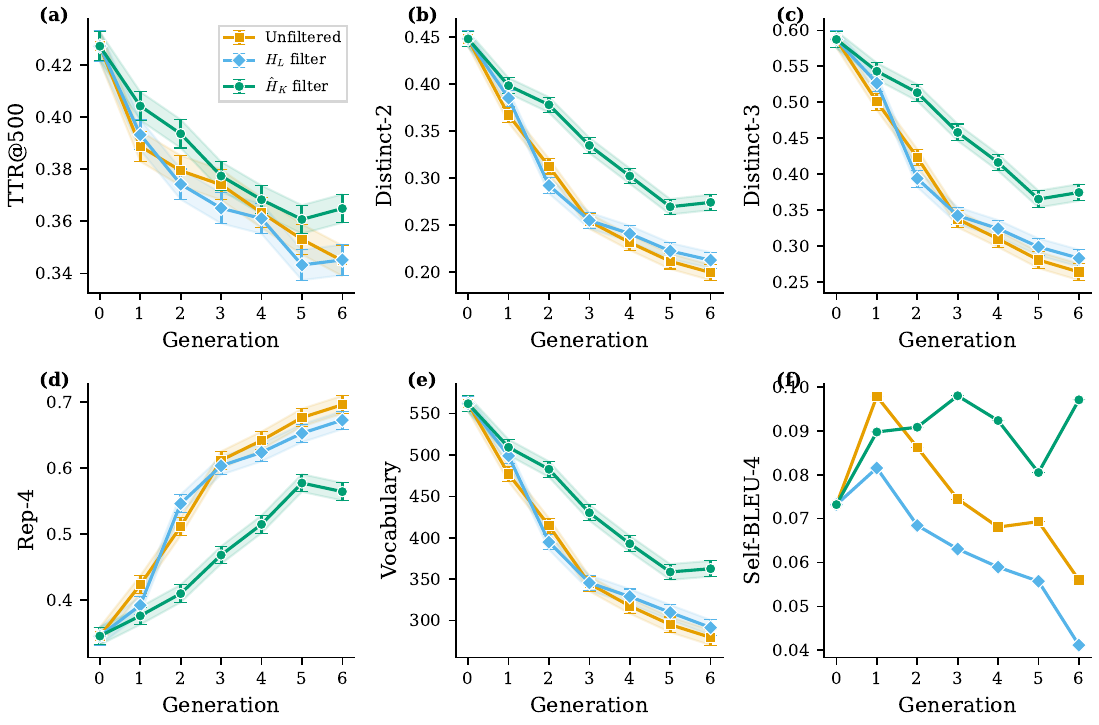}
  \caption{%
    Six text-diversity metrics across generations under the three conditions.
    $\hk$-filtering (green) maintains higher diversity on all within-document
    metrics (TTR, Distinct-$n$, vocabulary, Rep-4). Self-BLEU-4 (cross-document
    homogeneity) shows a non-monotonic pattern in early generations.
  }
  \label{fig:supp_text_metrics}
\end{figure}

\begin{figure}[ht]
  \centering
  \includegraphics[width=0.8\textwidth]{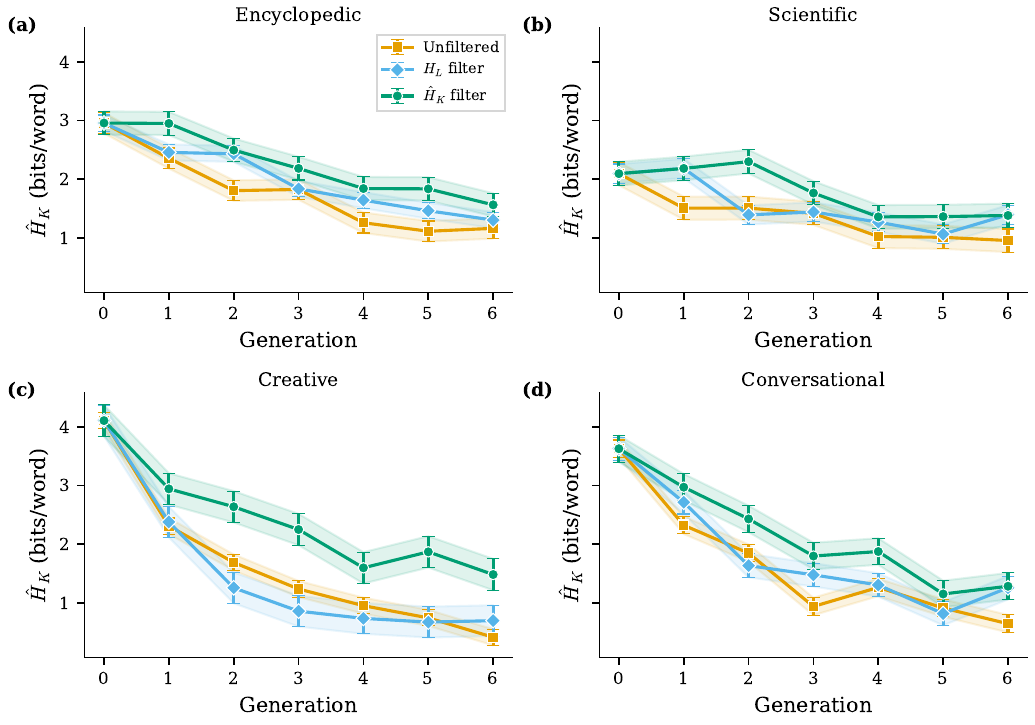}
  \caption{%
    Domain-level $\hk$ trajectories under the three conditions. The
    $\hk$-filter advantage is consistent across all four domains.
  }
  \label{fig:supp_domain}
\end{figure}

\begin{figure}[ht]
  \centering
  \includegraphics[width=0.7\textwidth]{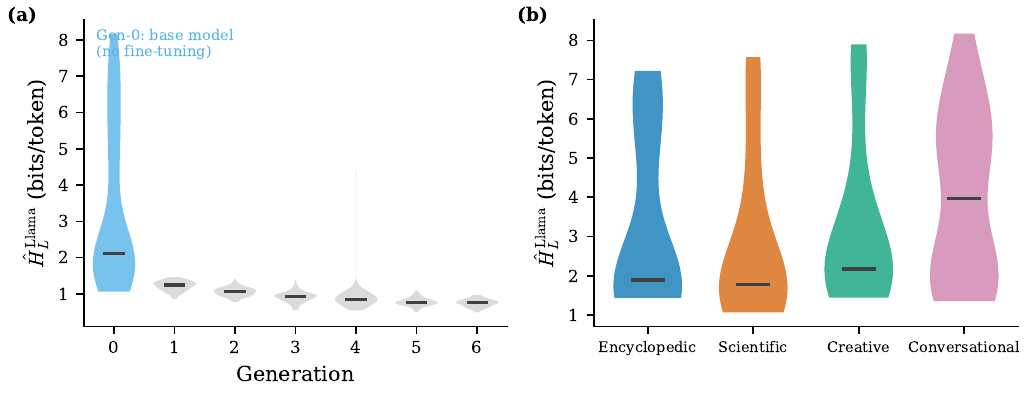}
  \caption{%
    Generation-0 $\hl$ distribution. The base Llama-3.1-8B-Instruct model
    exhibits a wide, bimodal $\hl$ distribution (mean~$=3.36$, std~$=2.17$),
    driven by topic-dependent familiarity. After one fine-tuning step the
    distribution collapses to a narrow regime (std~$\approx 0.13$), eliminating
    the variance the $\hl$-filter relies on.
  }
  \label{fig:supp_gen0}
\end{figure}

This appendix provides additional figures supporting the main experimental results.
The first two figures (Figures~\ref{fig:supp_three_metrics} and~\ref{fig:supp_text_metrics})
expand the quantitative picture of Section~\ref{sec:phase5} across multiple entropy
estimators and text-diversity metrics. Figure~\ref{fig:supp_domain} examines whether
the filtering advantage is uniform across the four document domains or concentrated
in particular content types.
Figure~\ref{fig:supp_gen0} provides a mechanistic explanation for the failure of
$\hl$-based filtering by characterising the generation-0 $\hl$ distribution.

\paragraph{Three-estimator comparison (Figure~\ref{fig:supp_three_metrics}).}
The main text reports results for $\hk$ as the primary collapse metric and briefly
notes the $\hl$ trajectories. Figure~\ref{fig:supp_three_metrics} places all three
estimators -- $\hk$, $\hl^{\text{Llama}}$, and the independent $\hl^{\text{GPT-2}}$
scorer -- side-by-side across generations, making two patterns visible simultaneously.
First, all three estimators agree on direction: the $\hk$-filter condition retains
higher entropy than $\hl$-filter, which in turn exceeds unfiltered, at generation~6.
Second, the $\hl^{\text{Llama}}$ panel shows the convergence that disables logprob
filtering: after generation~0, the three condition means are nearly indistinguishable
and their confidence bands overlap almost entirely, so a filter ranking documents by
$\hl$ within a generation operates on noise. The $\hk$ panel exhibits a qualitatively
different trajectory, with the $\hk$-filter condition maintaining a visible separation
from unfiltered throughout all six generations.

\paragraph{Full text-diversity metric suite (Figure~\ref{fig:supp_text_metrics}).}
Table~\ref{tab:phase5} in the main text reports four summary metrics at generation~6.
Figure~\ref{fig:supp_text_metrics} shows trajectories for six metrics across all
generations, providing a more complete view of how diversity evolves over the course
of the experiment. The within-document metrics (type--token ratio, Distinct-2,
Distinct-3, vocabulary size, and Rep-4) all tell a consistent story: the $\hk$-filter
condition (green) separates from the other two conditions by generation~1 and
maintains a persistent advantage, while the $\hl$-filter condition (orange) tracks
close to unfiltered on every within-document measure throughout. The Self-BLEU-4
\citep{papineni2002bleu} panel (cross-document homogeneity) shows a more complex non-monotonic pattern in
early generations, consistent with the discussion in Section~\ref{sec:phase5}: the
$\hk$-filter selects a statistically coherent cluster of documents, which briefly
increases cross-document similarity. This does not fully resolve by
generation~6: $\hk$-filter's Self-BLEU-4 remains distinctly higher
(0.086) than both unfiltered (0.075) and $\hl$-filter (0.060, both
$p<0.0001$, Welch's $t$-test) -- a small, sustained cross-document
homogeneity cost that partly offsets $\hk$-filtering's much larger
within-document diversity gains reported in Table~\ref{tab:phase5}.

\paragraph{Domain-level breakdown (Figure~\ref{fig:supp_domain}).}
A natural concern is that the aggregate filtering advantage reported in
Table~\ref{tab:phase5} could be driven by one domain in which $\hk$ happens to be
a particularly good signal, while other domains show little or no benefit.
Figure~\ref{fig:supp_domain} plots $\hk$ trajectories separately for each of the
four domains -- encyclopedic, scientific, conversational, and creative. The
$\hk$-filter advantage is present and directionally consistent across all four,
though the magnitude varies: the gap is largest in the conversational domain (where
collapse manifests most severely as phrase-level repetition, the exact phenomenon
$\hk$ measures) and smallest in the creative domain. No domain shows the $\hl$-filter
outperforming unfiltered on $\hk$, confirming that the null result for
$\hl$-filtering is not an artefact of domain averaging.

\paragraph{Generation-0 $\hl$ distribution (Figure~\ref{fig:supp_gen0}).}
Section~\ref{sec:phase5} argues that $\hl$-filtering fails because the within-generation
$\hl$ variance collapses after a single fine-tuning step, removing the ranking
signal the filter depends on. Figure~\ref{fig:supp_gen0} documents the starting
state that makes this explanation complete. At generation~0 -- documents produced by
the base Llama-3.1-8B-Instruct model before any fine-tuning -- the $\hl$ distribution
is wide and bimodal (mean~$= 3.36$, SD~$= 2.17$). The bimodality reflects
topic-dependent model familiarity: the base model assigns low entropy to topics
well-represented in its pretraining data and high entropy to unfamiliar topics,
producing two distinct modes rather than a smooth unimodal spread. After a single
generation of fine-tuning on 40 documents, the distribution collapses to a narrow
unimodal regime (SD~$\approx 0.13$, a 17-fold reduction), where all topics are now
equally familiar to the fine-tuned model. From generation~1 onward, every document
receives a similar $\hl$ score, so the $\hl$-filter's ranking is determined by
noise rather than signal. The wide generation-0 distribution also explains why the
$\hl$-filter is uncorrelated with $\hk$ at generation~0 ($r = 0.18$, $p=0.11$,
Section~\ref{sec:proxy}): the bimodal $\hl$ variation captures topic familiarity,
not document diversity.

\end{document}